\documentclass[runningheads]{llncs}
\pdfoutput=1 
 
\usepackage{eccv}

\usepackage{eccvabbrv}
\usepackage{multirow}
\usepackage{graphicx}
\usepackage{booktabs}

\usepackage[accsupp]{axessibility}  

\usepackage[pagebackref,breaklinks,colorlinks]{hyperref}

\usepackage{orcidlink}

\begin{document}

\title{ConStyle v2: A Strong Prompter for All-in-One Image Restoration} 

\author{Dongqi Fan \and
Junhao Zhang \and Liang Chang}

\authorrunning{Fan et al.}

\institute{University of Electronic Science and Technology of China, Chengdu, CN
\\
\email{\{dongqifan, junhaozhang\}@std.uestc.edu.cn} \\ \email{liangchang@uestc.edu.cn} 
} 

\maketitle

\begin{abstract}
This paper introduces ConStyle v2, a strong plug-and-play prompter designed to output clean visual prompts and assist U-Net Image Restoration models in handling multiple degradations. The joint training process of IRConStyle, an Image Restoration framework consisting of ConStyle and a general restoration network, is divided into two stages: first, pre-training ConStyle alone, and then freezing its weights to guide the training of the general restoration network. Three improvements are proposed in the pre-training stage to train ConStyle: unsupervised pre-training, adding a pretext task (i.e. classification), and adopting knowledge distillation. Without bells and whistles, we can get ConStyle v2, a strong prompter for all-in-one Image Restoration, in less than two GPU days and doesn't require any fine-tuning. Extensive experiments on Restormer (transformer-based), NAFNet (CNN-based), MAXIM-1S (MLP-based), and a vanilla CNN network demonstrate that ConStyle v2 can enhance any U-Net style Image Restoration models to all-in-one Image Restoration models. Furthermore, models guided by the well-trained ConStyle v2 exhibit superior performance in some specific degradation compared to ConStyle. The code is avaliable at: \href{https://github.com/Dongqi-Fan/ConStyle_v2}{https://github.com/Dongqi-Fan/ConStyle\_v2}
\end{abstract}
\section{Introduction}
\label{sec:intro}
Image Restoration (IR) is a fundamental vision task in the computer vision community,  which aims to reconstruct a high-quality image from a degraded one. Recent advancements in deep learning have shown promising results in specific IR tasks such as denoising \cite{1cheng2021nbnet, 2chang2020spatial, 3yue2019variational}, dehazing \cite{4liu2019griddehazenet, 5ren2018gated, 6dong2020physics}, deraining \cite{7gu2024networks, 8ren2019progressive, 9jiang2020multi}, desnowing \cite{10chen2021all, 11chen2020jstasr, 12liu2018desnownet}, motion deblurring \cite{13kong2023efficient, 14cui2023dual, 15yang2022motion}, defocus deblurring \cite{16ruan2022learning, 17zhang2022dynamic, 18lee2021iterative}, low-light enhancement \cite{19wang2023low, 20cai2023retinexformer, 21fu2023you}, and JPEG artifact removal/correction \cite{64jiang2021towards, 65zheng2019implicit}. However, these models are limited to addressing only one specific type of degradation. To tackle this issue, researchers have focused on developing models capable of handling multiple degradations \cite{22zamir2021multi, 23cui2023irnext, 24wang2022uformer, 25li2023efficient}. Yet, these models require retraining for each different type of degradation, that is a set of weights is tailored for a single type of degradation. Obviously, these approaches are not practical as multiple degradations often coexist in real-world scenarios. For instance, rainy days are often associated with haze and reduced lighting.

\begin{figure}[tb]
\centering
\includegraphics[width=\linewidth]{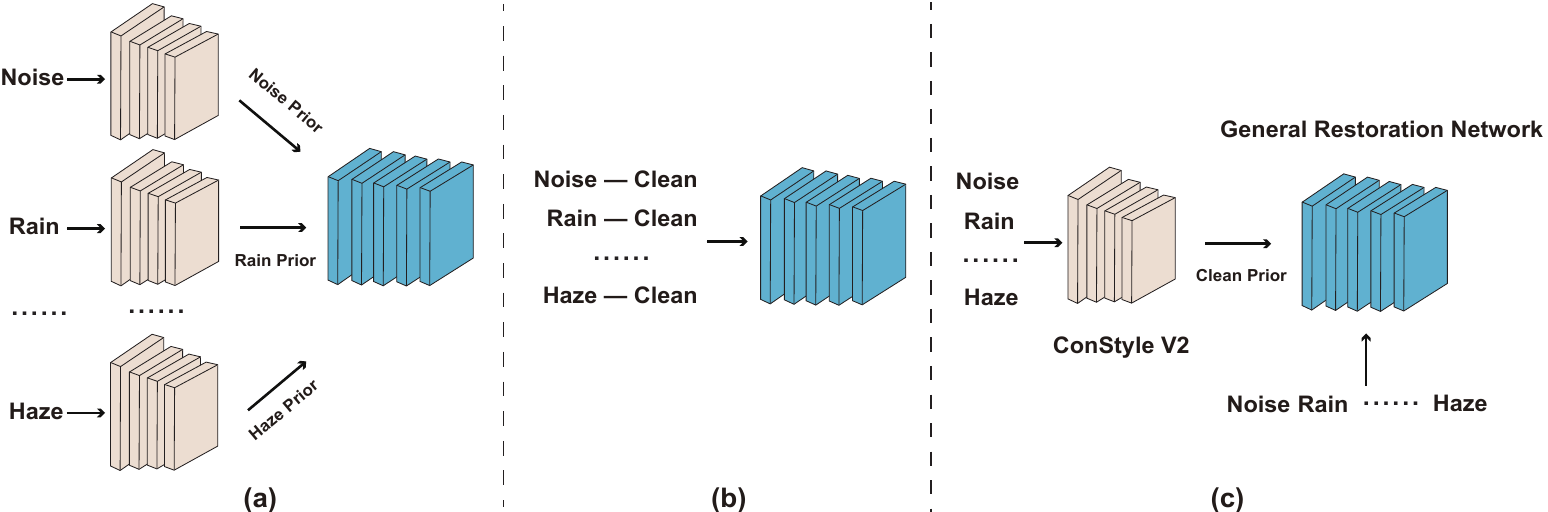}
\caption{Different way to solve multiple degradations. (a) The priors are obtained by setting sub-networks as many as degradations. (b) The example pair of the degradation-clean must provide in training and inference stage. (c) ConStyle v2 adaptively outputs clean visual prompts according to different degradations to guide the training of the general restoration network.}     
\label{fig:intro}
\end{figure}

The all-in-one Image Restoration \cite{26zhang2023ingredient, 27park2023all, 28valanarasu2022transweather, 29chen2022learning} is a kind of method that only uses a suit of weights to address multiple types of degradation. However, these all-in-one models often have a large number of parameters and require heavy computations, leading to time-consuming and inefficient training processes. For instance, Chen et al. \cite{29chen2022learning} (\cref{fig:intro} (a)) adjust the number of teacher networks based on the number of degradations. Thus, the more teacher networks there are, the more complex the training process becomes; Li et al. \cite{32li2020all} (\cref{fig:intro} (a)) also adopt the number of sub-networks the same as the amount of degradation, and its backbone is obtained through neural architecture search (NAS); PromptGIP \cite{31liu2023unifying} (\cref{fig:intro} (b)) leverage the idea of visual prompting to help the model in handling multiple degradations, but a degradation-clean sample pair must be provided in the training and inference stage and requiring 8 V100 GPUs for training. These methods are inefficient due to the lack of prior knowledge about the degradations in the input image.  In other words, the prior information about the degradation type needs to be first obtained and then passed to subsequent sub-networks. In contrast, thanks to ConStyle v2, general restoration network (\cref{fig:intro} (c)) do not need specific prior knowledge about degradations, but rather require a clean visual prompt (clean prior).

\begin{figure}[tb]
\centering
\includegraphics[width=\linewidth]{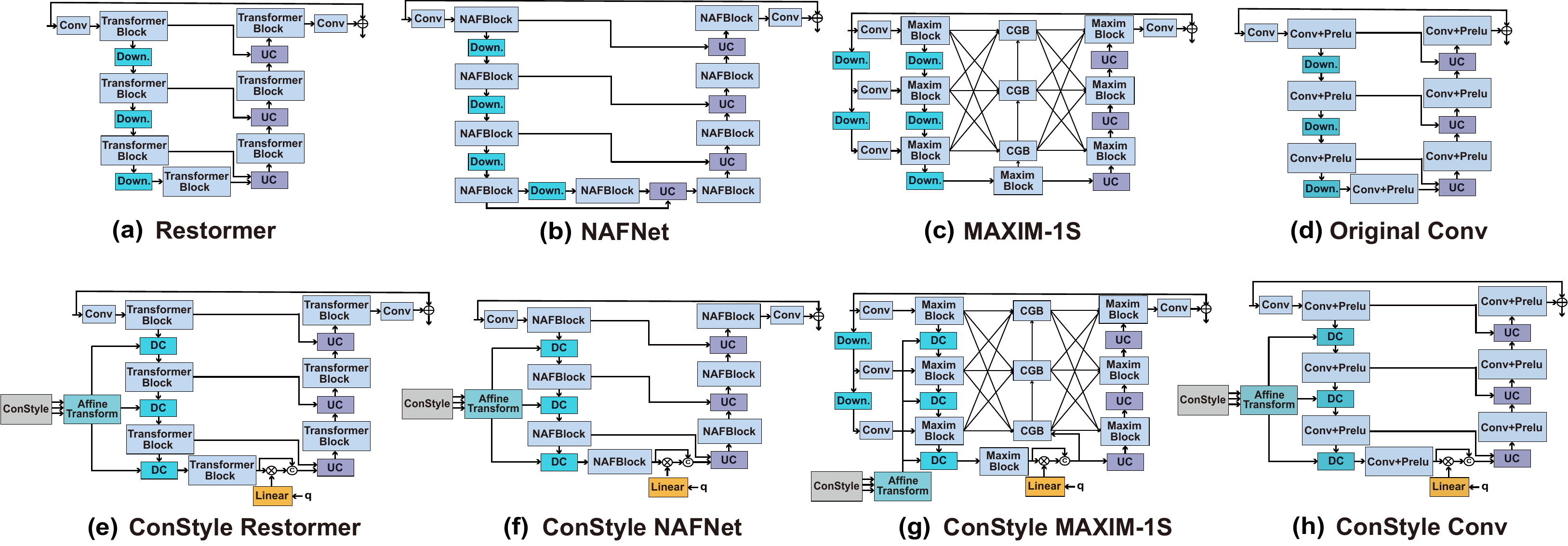}
\caption{The detailed structure of the original models (a)(b)(c)(d) and the ConStyle/ConStyle v2 models (e)(f)(g)(h). DC represents the downsample and
concat operation, and UC represents upsample and concat operation}     
\label{fig:all_models}
\end{figure}

In this paper, we introduce a strong prompter for all-in-one Image Restoration: ConStyle v2 (\cref{fig:constyle_v2} and \cref{fig:difference}). The key of our work is to ensure that ConStyle v2 generates clean visual prompts, thus mitigating the issue of model collapse and guiding the training of general restoration networks. Model collapse typically arises when a model struggles to simultaneously handle multiple degradations. To address this challenge, we propose three simple yet effective improvements to ConStyle: unsupervised pre-training, leveraging a pretext task to enhance semantic information extraction capabilities, and employing knowledge distillation to further enhance this capacity. ConStyle v2 consists of convolution, linear, and BN layers and without any complex operators (\cref{fig:difference} (c)), so the time of the training is less than two days on a V100 GPU and Intel Xeon Silver 4216 CPU. Once trained, ConStyle v2 can be seamlessly integrated into any U-Net model to facilitate their training without fine-tuning. Additionally, given the lack of datasets encompassing multiple degradations, we collect and produce a Mix Degradations dataset, which includes noise, motion blur, defocus blur, rain, snow, low-light, JPEG artifact, and haze, to cater to training requirements. To verify our method, we perform ConStyle v2 on three state-of-the-art IR models (Restormer\cite{36zamir2022restormer}, MAXIM-1S\cite{37tu2022maxim}, NAFNet\cite{38chen2022simple}) and a non-IR U-Net model consisting of vanilla convolutions. \cref{fig:all_models} shows details architecture of Original models and ConStyle/ConStyle v2 models. Experiment results on 27 benchmarks demonstrate that our ConStyle v2 is a powerful plug-and-play prompter for all-in-one image restoration and exhibits superior performance for specific degradation compared to ConStyle \cite{fan2024irconstyle}.

Our contributions can be summarized as follows:
\begin{itemize}
\item Three simple yet effective enhancements are proposed to train ConStyle v2, and the time of the training is less than two GPU days.
\item We propose a Mix Degradations dataset, which includes noise, motion blur, defocus blur, rain, snow, low-light, JPEG artifact, and haze, to cater to training needs.
\item We propose a strong plug-and-play prompter for all-in-one and specific image restoration, in which the model collapse issue is avoided. 
\end{itemize}
\section{Related Work}
\label{sec:related}

\subsection{All-in-One Image Restoration}
While numerous works \cite{22zamir2021multi, 23cui2023irnext, 24wang2022uformer, 25li2023efficient, 36zamir2022restormer, 37tu2022maxim, 38chen2022simple, 39chen2021pre, 40cui2022selective, 41li2022drcnet, 56fan2024lir} excel in various Image Restoration tasks, they are typically limited to addressing a single type of degradation with a specific set of weights. To solve this problem, all-in-one Image Restoration (IR) methods \cite{26zhang2023ingredient, 27park2023all, 28valanarasu2022transweather, 29chen2022learning, 30luo2023controlling, 31liu2023unifying, 33park2023all, 34zhang2023ingredient, 42li2022all, 43ma2023prores, 44potlapalli2023promptir} have been developed. These methods aim to enable models to effectively handle multiple degradations simultaneously. For example, AirNet \cite{42li2022all} leverages MoCo \cite{46he2020momentum} and Deformable Convolution \cite{45dai2017deformable} to transform degradation priors obtained from the former into convolution kernels in the latter, enabling dynamic degradation removal; DA-CLIP \cite{30luo2023controlling} builds upon the architecture of CLIP \cite{47radford2021learning}, in which BLIP \cite{48li2022blip} is used to generate synthetic captions for high-quality images. Then match low-quality images with captions and corresponding degradation types as image-text-degradation pairs; ADMS \cite{33park2023all} introduces a Filter Attribution method based on FAIG \cite{35xie2021finding} to identify the specific contributions of filters in removing specific degradations, while IDR \cite{34zhang2023ingredient} proposes a learnable Principal Component Analysis and treats various IR tasks as a form of multi-task learning to acquire priors. Different from the above methods, we aim to design a plug-and-play module that can transform a non-all-in-one model into an all-in-one model.

\subsection{Visual Prompting}
In the field of Natural Language Processing (NLP), Prompting Learning refers to providing task-specific instructions or in-context information to a model without the need for retraining. This approach has shown promising results in NLP, such as GPT-3 \cite{49brown2020language}. Drawing inspiration from Prompting Learning in NLP, recently, there have been many excellent visual prompting works in the IR \cite{31liu2023unifying, 43ma2023prores, 44potlapalli2023promptir, 50chen2022unified, 51liu2023explicit, 52bar2022visual, 53wang2023images, 54shen2023aligning}. For example, ProRes \cite{43ma2023prores} involves adding a target visual prompt to an input image to create a "prompted image". This prompted image is then flattened into patches, with the weights of ProRes frozen, and learnable prompts are randomly initialized for new tasks or datasets; PromptIR \cite{44potlapalli2023promptir}  introduces a Prompt Block in the decoder stage of the U-Net architecture. This block takes prompt components and the output of the previous transformer block as inputs, with its output being fed into the next transformer block;  PromptGIP \cite{31liu2023unifying} proposes a training method akin to masked autoencoding, where certain portions of question images and answer images are randomly masked to prompt the model to reconstruct these patches from the unmasked areas. During inference, input-output pairs are assembled as task prompts to realize image restoration. Our approach also leverages visual prompting, but in a more efficient manner, eliminating the need to distinguish different degradations like the above methods. It will provide a clean visual prompt for other models.
\begin{figure}[tb]
\centering
\includegraphics[width=0.95\linewidth]{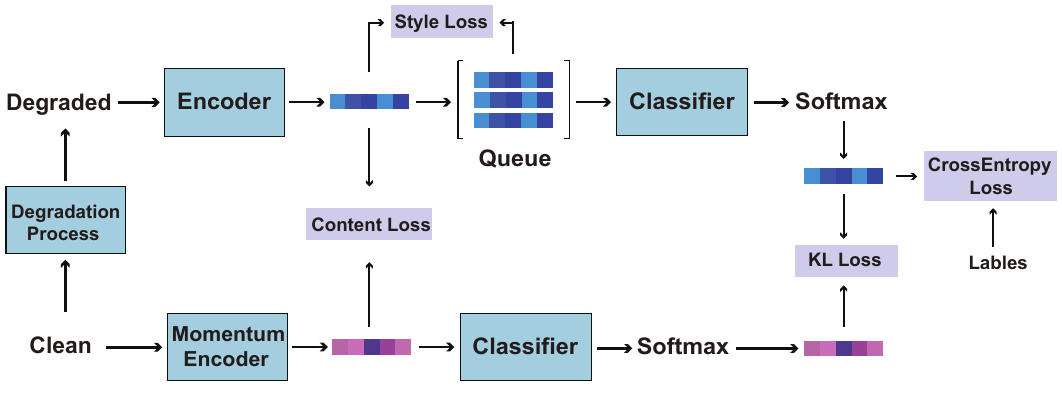}
\caption{The training diagram of the ConStyle v2. Only the Encoder is retained once training is complete.}     
\label{fig:constyle_v2}
\end{figure}

\begin{figure}[t]
\centering
\includegraphics[width=\linewidth]{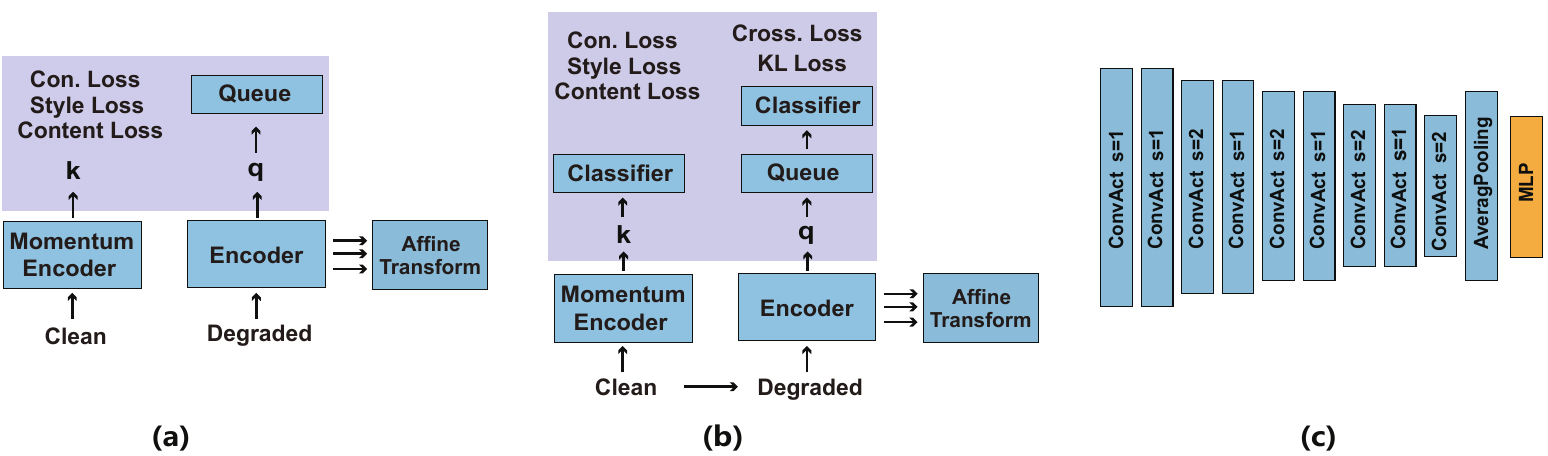}
\caption{The difference between ConStyle (a) and ConStyle v2 (b), and the detail structure of the Momentum Encoder and Encoder (c). Where Con. Loss and Cross. Loss are abbreviations of Content Loss and CrossEntropy Loss. In ConStyle, the Momentum Encoder and Queue are only removed in the inference stage, while, in ConStyle v2, they are removed when the pre-training is finished.}     
\label{fig:difference}
\end{figure}

\section{Method}
\label{sec:method}
The training diagram of ConStyle v2 is depicted in \cref{fig:constyle_v2}, while the distinctions between ConStyle and ConStyle v2 are illustrated in \cref{fig:difference}. The same as ConStyle, ConStyle v2 only retains the Encoder part when the pre-training is complete. In this section, we first provide a brief overview of ConStyle (\cref{sec:review}), followed by showing problems encountered with ConStyle in multiple degradations (\cref{sec:problem}), and finally, we illustrate the improvements made from ConStyle to ConStyle v2 in three steps (\cref{sec:constylev2}). In addition, the Mix Degradations dataset is described in \cref{sec:datasets}.

\subsection{Review of ConStyle}
\label{sec:review}
IRConStyle \cite{fan2024irconstyle} is a versatile and robust IR framework consisting of the ConStyle and a general restoration network. ConStyle includes several convolutional layers and one MLP layer, which is responsible for extracting latent features (the latent code and intermediate feature map) and then passing them to the general restoration network. The general restoration network follows an abstract U-Net style architecture, allowing for the instantiation of any IR U-Net model. The training stage (\cref{equ1}) and inference (\cref{equ2}) process of IRConStyle \cite{fan2024irconstyle} can be described as follows:

\begin{equation}
I_{restored}=G(E(I_{degraded}, I_{clean}), I_{degraded})
\label{equ1}
\end{equation}
\begin{equation}
I_{restored}=G(E(I_{degraded}), I_{degraded})
\label{equ2}
\end{equation}

Where G stands for general restoration network, E for ConStyle, $I_{degraded}$ for the input degraded image, and $I_{restored}$ for the output restored clean image. Based on the contrast learning framework MoCo \cite{46he2020momentum}, ConStyle cleverly integrates the idea of style transfer and replaces the pretext task, Instance Discrimination \cite{55wu2018unsupervised}, with one pretext task more suitable for IR. The total loss functions for IRConStyle are as follows:

\begin{equation}
L_{total}=L_{style}+L_{content}+L_{InfoNCE}+L_1
\label{equ3}
\end{equation}

The calculation of $L_{style}$, $L_{content}$, and $L_{InfoNCE}$ is performed in ConStyle, while $L_1$ is performed in general restoration network. Under the supervision of $L_{style}$, $L_{content}$, and $L_{InfoNCE}$, latent features move closer to the clean space and further away from the degradation space. Since ConStyle can adaptively output clean latent features according to input degraded images, it is natural for us to believe that ConStyle should be able to turn the general restoration network into an all-in-one model. However, the experiment results on ConStyle models are not as expected.

\subsection{Mix Degradations Datasets}
\label{sec:datasets}
We need a training dataset that includes noise, motion blur, defocus blur, rain, snow, low light, JPEG artifact, and haze, but the existing training dataset did not meet our needs. Therefore, we propose a dataset, namely Mix Degradations datasets, consisting of image pairs with all of the aforementioned degradations. Details on the Mix Degradations dataset can be found in \cref{datasets}. The images with noise and JPEG artifacts are respectively generated using established methods same as \cite{36zamir2022restormer, 37tu2022maxim, 38chen2022simple} and \cite{63liang2021swinir, 64jiang2021towards, 65zheng2019implicit}. It is important to note that in the OTS dataset, haze images with intensities of 0.04 and 0.06 are manually removed due to being too clear to the human eye. In addition, the deraining dataset, which includes Rain14000 \cite{rain14000fu2017removing}, Rain1800 \cite{rain1800yang2017deep}, Rain800 \cite{rain800zhang2019image}, and Rain12 \cite{rain12li2016rain}, initially contained 13,712 images, but two erroneous pictures are identified and removed. After a unified cropping process, the Mix Degradations dataset has 621,573 images, all of size 256 × 256. The Mix Degradations datasets, the uncropped joint datasets mentioned in \cref{datasets} (totaling 46,301 images), and the data preparation file are all available on our GitHub link.

\begin{table}[tb]
\caption{Details about Mix Degradations datasets.}
\scalebox{0.75}{
\begin{tabular}{ccccc}
\toprule[1.3pt]
Task         & Motion Blurring       & Defocus Blurring & Dehazing     & Low-light Enhacement        \\ \midrule[0.8pt]
Where        & GoPro \cite{gopro}                 & LFDOF \cite{lfdofruan2021aifnet}            & OTS \cite{ots_sots} & LoL v1 \cite{lolv1wei1808deep}, LoL v2 \cite{lolv2yang2021sparse}, and FiveK \cite{fivekbychkovsky2011learning}   \\
Used Number  & 2,103                 & 5,606            & 12,500       & 5,885                       \\
Crop Size    & 256                   & 256              & 256          & 256                         \\
Step Size    & 150                   & 220              & 240          & 100                         \\
Final Number & 84,120                & 84,090           & 82,425       & 51,891                      \\ \midrule[0.8pt]
Task         & JPEG Artifact Removal & Denoising        & Desnowing    & Deraining                   \\ \midrule[0.8pt]
Where        & DIV2K \cite{div2k}                 & DIV2K \cite{div2k}           & Snow100K \cite{12liu2018desnownet}     & Rain14000, 1800, 800 and 12 \\
Used Number  & 800                   & 800              & 7000         & 13710                       \\
Crop Size    & 256                   & 256              & 256          & 256                         \\
Step Size    & 165                   & 165              & 160          & 240                         \\
Final Number & 77,904                & 77,904           & 82,432       & 80,807                      \\ \bottomrule[1.3pt]
\end{tabular}}
\label{datasets}
\end{table}

\subsection{ConStyle on Mix Degradations Datasets}
\label{sec:problem}
To evaluate whether ConStyle \cite{fan2024irconstyle} can directly convert U-Net models to all-in-one models, we conduct experiments using Original models (Restormer \cite{36zamir2022restormer}, NAFNet \cite{lfdofruan2021aifnet}, MAXIM-1S \cite{37tu2022maxim}) and ConStyle models (ConStyle Restormer, ConStyle NAFNet, ConStyle MAXIM-1S) on Mix Degradations datasets. These models will be tested on GoPro \cite{gopro} (motion blurring), RealDOF \cite{lfdofruan2021aifnet} (defocus blurring), LoL v1 \cite{lolv1wei1808deep} (low-light enhancement), SOTS outdoors \cite{ots_sots} (dehazing), LIVE1 \cite{livesheikh2005live} (JPEG artifact removal), CSD \cite{10chen2021all} (desnowing), CBSD68 \cite{bsd68} (denoising), and Rain100H \cite{rain100l} (deraining). In addition, we introduce two more models for comparison: Original Conv and ConStyle Conv. The Original Conv is a vanilla U-Net convolution model, while the ConStyle Conv incorporates ConStyle into the Original Conv. These two additional models are used to verify the generality of the ConStyle and ConStyle v2. It is important to note that to expedite evaluation during the training, only a subset of the test datasets is used. For example, only 24 images from GoPro's test dataset of 1,111 images are selected. Using the full test datasets for all 8 tasks would significantly increase training time, as inference on Restormer alone with GoPro's test dataset would take 40 minutes on a V100 GPU.

\begin{figure}[tb]
\centering
\includegraphics[width=\linewidth]{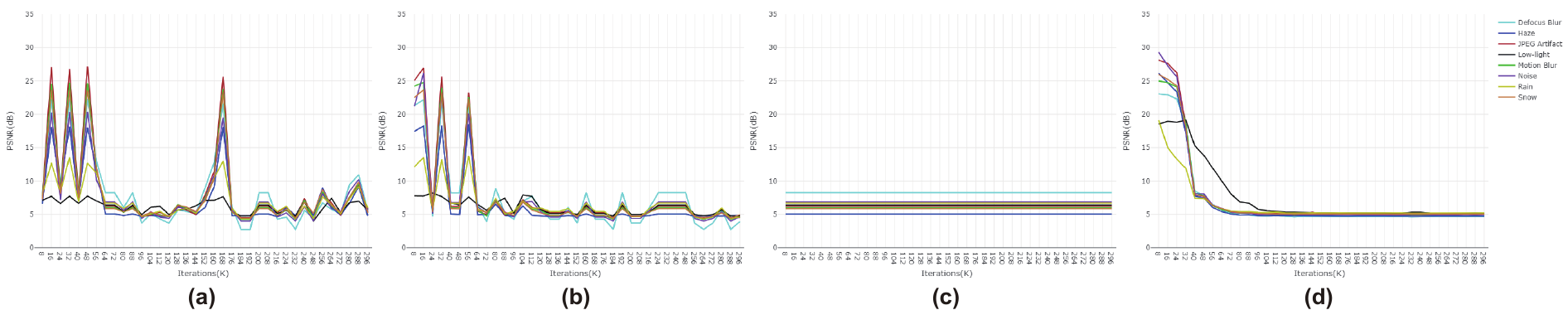}
\caption{Original Conv (a), ConStyle Conv (b), Restormer (c), and ConStyle Restormer (d) trained on Mix Degradations datasets.}     
\label{fig:origin_constyle}
\end{figure}

The results in \cref{fig:origin_constyle} demonstrate that under multiple degradation settings, the performance of ConStyle Conv not only fails to surpass that of the Original Conv but also remains consistently low performance after 80K iterations. While ConStyle Restormer shows better performance than Restormer, it also suffers from model collapse problems as Restormer. In the following section, we will illustrate the improving process of ConStyle to ConStyle v2 on Restormer and Original Conv step by step.

\subsection{ConStyle v2}
\label{sec:constylev2}
\subsubsection{Unsupervised Pre-training}
We find that even for specific IR tasks, ConStyle models exhibit varying degrees of model collapse issues. For example, in the dehazing task, the performance of ConStyle NAFNet, ConStyle MAXIM-1S, and ConStyle Restormer significantly declines after 250K iterations, 30K iterations, and 10K iterations respectively. Interestingly, even within the same model like ConStyle MAXIM-1S, the onset of model collapse differs across denoising, deraining, and deblurring tasks, occurring at 10K, 100K, and 200K iterations respectively. To address this challenge, IRConStyle \cite{fan2024irconstyle} implements a strategy of early stopping ConStyle updates. Here, we intend to elegantly solve this problem. Since this problem happens in joint training, then it is natural to split joint training process into two stages. Specifically, ConStyle is pre-trained independently, followed by fixing its weights and integrating it with other IR models for guided training. For pre-training stage, we leverage the generation techniques of Real-ESRGAN \cite{57wang2021real} and ImageNet-C \cite{58hendrycks2019benchmarking} in the Degradation Process (\cref{fig:constyle_v2}) for unsupervised training on ImageNet-1K \cite{59deng2009imagenet}.

\begin{figure}[tb]
\centering
\includegraphics[width=\linewidth]{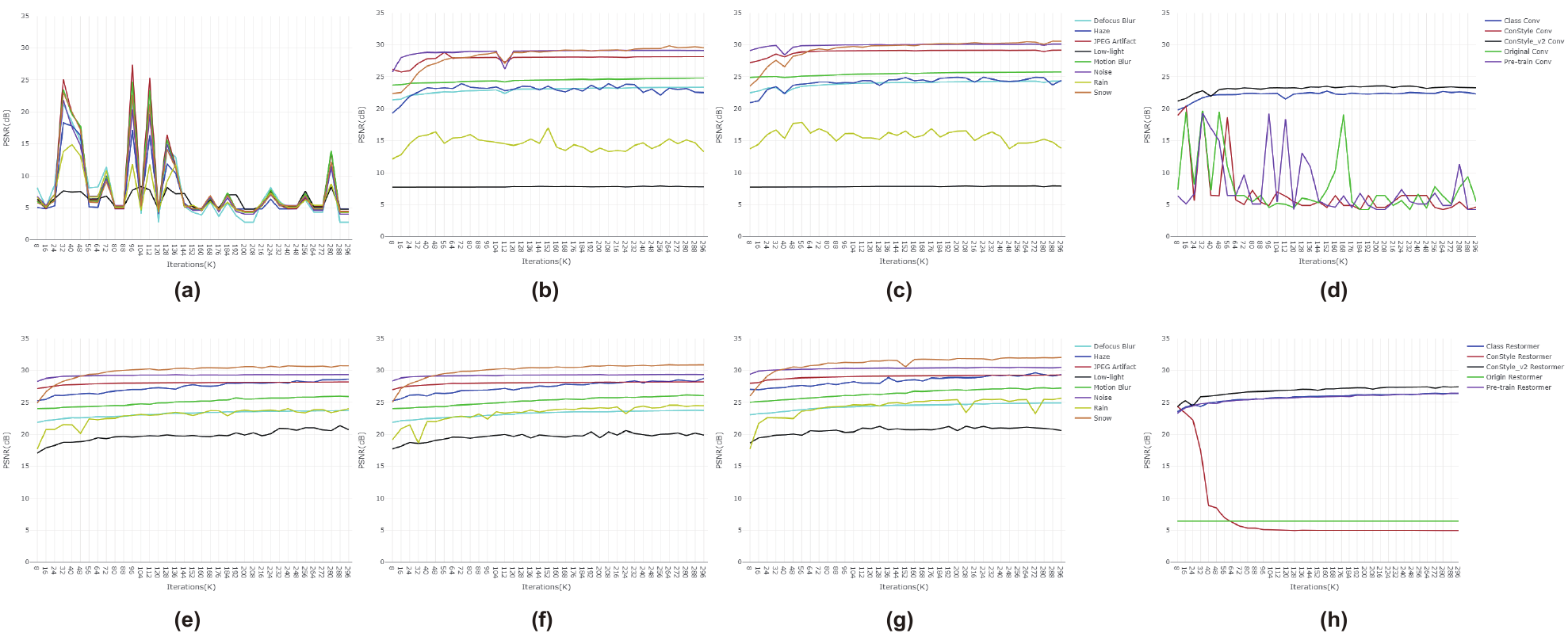}
\caption{(a), (b), and (c) represent Pre-train ConStyle Conv, Class ConStyle Conv, and ConStyle v2 Conv. (e), (f), and (g) represent Pre-train ConStyle Restormer, Class ConStyle Restormer, and ConStyle v2 Restormer. (d) and (h) represent average performance on eight degradations.}     
\label{fig:base_class}
\end{figure}

Since our goal is to train ConStyle v2 to be a powerful prompter that can produce a clean visual prompt based on the different degradations, we use the method in ImageNet-C \cite{58hendrycks2019benchmarking} to generate motion blur, snow, and low contrast and the two-stage degradation method in Real-ESRGAN \cite{57wang2021real} to generate Gaussian blur, noise, and JPEG artifacts. For each batch of images, 40\% of the images are randomly selected to add motion blur and snow and change contrast, while 60\% of the images are added Gaussian blur, noise, and JPEG artifact. For details training setting of the pre-training please see \cref{sec:experiments}. The pre-trained ConStyle, with the weight fixed, is incorporated into the general restoration network for training on the Mix Degradations datasets. Here, we name the models of this stage as Pre-train models. The results of Pre-train Restormer and the Pre-train Conv can be seen in \cref{fig:base_class} (a) and (e).

\subsubsection{Pretext Task}

\begin{figure}[tb]
\centering
\includegraphics[width=\linewidth]{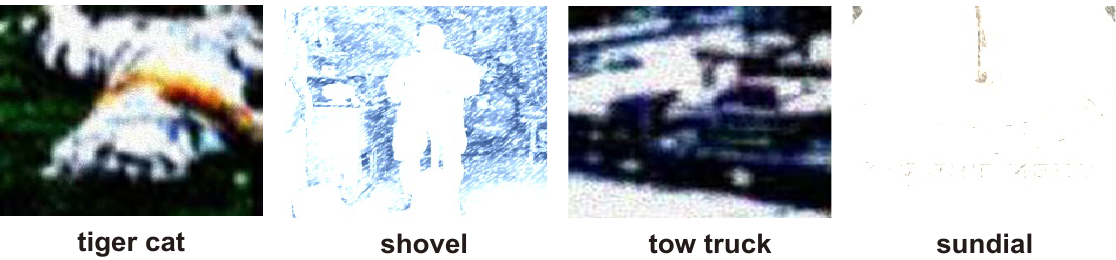}
\caption{The visual results of the proposed degradation process. Several degradations are added to a single image, and the intensity of the degradations is random.}     
\label{fig:generation}
\end{figure}

Following pre-training step, ConStyle Restormer has demonstrated significant performance improvement, successfully resolving the issue of model collapse. Conversely, ConStyle Conv continues to face challenges with unstable training and limited enhancement in performance. We believe that this is attributed to heavy degradation, leading to a loss of semantic information (\cref{fig:generation}) in the original image. It makes the model of weak semantic extraction ability, such as ConStyle Conv, also have poor image restoration performance. Because the process of image restoration involves pixel-wise operations and necessitates a comprehensive understanding of the entire image. Thus we introduce a pretext task (classification) to enhance the semantic information extraction capabilities of ConStyle, so as to improve such capability of ConStyle models. Specifically, we add Classifier and Softmax layers at the back of the Encoder and leverage the labels of the ImageNet (\cref{fig:constyle_v2}). Here, we name the models of this stage as Class models. As shown in \cref{fig:base_class} (b) and (f), the addition of the pretext task has little influence on ConStyle Restormer, since the Transformer model already has strong semantic extraction abilities. In contrast, for ConStyle Conv, the inclusion of the pretext task makes the training stable, and the performance is significantly improved.

\subsubsection{Knowledge Distillation} 
Although ConStyle has been significantly improved through pre-training and the addition of a pretext task, enabling it to generate clean visual prompts based on degraded image input, to further boost ConStyle's ability to extract semantic information, we take the last step to improve ConStyle to ConStyle v2. Inspired by BYOL\cite{60grill2020bootstrap}, SimSam\cite{61chen2021exploring}, and DINO\cite{62caron2021emerging}, we take advantage of knowledge distillation. Since the input of the Momentum Encoder is the clean image, and its output visual prompt is cleaner than the output of Encoder, by utilizing the Momentum Encoder as a teacher and the Encoder as a student, the teacher network is able to adaptively guide the student network during training. Specifically, a Classifier and Softmax layer are added at the back of the Momentum Encoder, with distance measured by the Kullback-Leibler (KL) function (\cref{fig:constyle_v2}). Now, we have the final ConStyle v2, and the performance of ConStyle v2 Restormer (\cref{fig:base_class} (g) and ConStyle v2 Conv (\cref{fig:base_class} (c)) is raised again compared with other models. In addition, each improvement in the average performance across eight degradations is depicted in \cref{fig:base_class} (d) and (h).

\section{Experiments}
\label{sec:experiments}
\subsection{Implement details}
All experiments in this paper are performed on an NVIDIA Tesla V100 GPU. To be consistent with ConStyle \cite{fan2024irconstyle}, we use AdamW ($\beta_{1}$=0.9, $\beta_{2}$=0.999, weight decay=$1e^{-4}$) optimizer with an initial learning rate of $3e^{-4}$ and Cosine annealing. \textbf{Training Stage:} The batch size, crop size, and total iterations are set as 16, 128, and 700K respectively. \textbf{Pre-training Stage:} The batch size, crop size, and total iterations are set as 32, 224, and 200K respectively. In the process of generating degraded images, we directly use all configurations in Real-ESRGAN \cite{57wang2021real} and change the intensity of degradation in ImageNet-C \cite{58hendrycks2019benchmarking}.

\begin{table}[tb]
\centering
\caption{Parameters, computations, and inference speed comparison. $(*)$ represents ConStyle, Pre-train, class, and ConStyle v2 models. For instance, Conv$(*)$ stands for ConStyle Conv, Pre-train Conv, Class Conv, and ConStyle v2 Conv.}
\scalebox{0.95}{
\begin{tabular}{ccccc}
\toprule[1.3pt]
Method    & \makebox[0.19\textwidth][c]{Restormer \cite{36zamir2022restormer}} &  \makebox[0.19\textwidth][c]{Restormer$(*)$} & \makebox[0.17\textwidth][c]{NAFNet \cite{38chen2022simple}} &  \makebox[0.16\textwidth][c]{NAFNet$(*)$} \\ \midrule[0.8pt]
Params(M) & 26.12     & 15.57              & 87.40         & 12.74                                \\
GFLOPs    & 70.49     & 74.92              & 49.13         & 46.97                                \\
Speed(us) & 60        & 61                 & 53            & 54                                   \\ \midrule[0.8pt]
Method    & MAXIM-1S \cite{37tu2022maxim}     &  MAXIM-1S$(*)$     & Conv & Conv$(*)$                        \\ \midrule[0.8pt]
Params(M) & 8.17      & 8.10               & 5.03          & 6.78                                 \\
GFLOPs    & 21.58     & 25.58              & 19.64         & 25.90                                \\
Speed(us) & 53        & 52                 & 3             & 5                                    \\ \bottomrule[1.3pt]
\end{tabular}}
\label{comparaision}
\end{table}

\begin{table}[tb]
\centering
\caption{The average performance of the PSNR/SSIM on 27 benchmarks. Red means the best and blue means the second best.}
\scalebox{0.9}{
\begin{tabular}{cccccc}
\toprule[1.3pt]
          & \makebox[0.17\textwidth][c]{Original}     & \makebox[0.17\textwidth][c]{ConStyle}     & \makebox[0.17\textwidth][c]{Pre-train}     & \makebox[0.17\textwidth][c]{Class}        & \makebox[0.17\textwidth][c]{ConStyle v2 }   \\ \midrule[0.8pt]
Restormer \cite{36zamir2022restormer} & 6.40/0.0506   & 25.23/0.7844 & 27.24/0.8222 & {\color[HTML]{3166FF} 27.40/0.8305} & {\color[HTML]{FE0000} 27.60/0.8352} \\
NAFNet \cite{38chen2022simple}    & 27.45/0.8347 & 26.57/0.8054 & 27.43/0.8341 & {\color[HTML]{3166FF} 27.53/0.8379} & {\color[HTML]{FE0000} 27.55/0.8391} \\
MAXIM-1S \cite{37tu2022maxim}  & 26.95/0.8319 & 26.63/0.8290 & 27.14/0.8319 & {\color[HTML]{3166FF} 27.41/0.8372} & {\color[HTML]{FE0000} 27.54/0.8400} \\
Conv      & 21.44/0.6177 & 20.99/0.5986 & 20.14/0.6544 & {\color[HTML]{3166FF} 24.84/0.7889} & {\color[HTML]{FE0000} 26.06/0.7974} \\ \bottomrule[1.3pt]
\end{tabular}}
\label{averge_performance}
\end{table}

Mix Degradations datasets is used in the training stage and ImageNet-1K in the pre-training stage. For evaluation, GoPro \cite{gopro}, HIDE \cite{hide}, RealBlur-J \cite{realblur}, and RealBlur-R \cite{realblur} are used for motion deblurring, DPDD \cite{dpddabuolaim2020defocus} is used for defocus deblurring, SOTS outdoors \cite{ots_sots} is used for dehazing, Rain100H \cite{rain100l}, Rain100L \cite{rain100l}, Test1200 \cite{test1200zhang2018density}, and Test2800 \cite{test2800fu2017removing} are used for deraining, FiveK \cite{fivekbychkovsky2011learning}, LoL v1 \cite{lolv1wei1808deep}, and LoL v2 \cite{lolv2yang2021sparse} are used for low-light enhancement, CSD \cite{10chen2021all}, Snow100K (S, M, and L) \cite{12liu2018desnownet} are used for desnowing, CBSD68 \cite{bsd68} and urban100\cite{urban100} are used for denoising, and LIVE1 \cite{livesheikh2005live} is used for JPEG artifact removal.

\begin{table}[tb]
\caption{RGB image denoising tested by PSNR. Where ($*$) indicates models are trained with random sigma (0 to 50), while other indicates models are trained with fixed sigma. Blue means better and red means worse.}
\centering
\scalebox{0.8}{
\begin{tabular}{ccccccccccccc}
\toprule[1.3pt]
\multirow{2}{*}{Method} & \multicolumn{3}{c}{CBSD68(*) \cite{bsd68}} & \multicolumn{3}{c}{CBSD68 \cite{bsd68}} & \multicolumn{3}{c}{Urban100(*) \cite{urban100}} & \multicolumn{3}{c}{Urban 100 \cite{urban100}} \\ \cmidrule{2-13} 
                        & $\sigma$=15       & $\sigma$=25       & $\sigma$=50      & $\sigma$=15      & $\sigma$=25      & $\sigma$=50     & $\sigma$=15        & $\sigma$=25       & $\sigma$=50       & $\sigma$=15       & $\sigma$=25       & $\sigma$=50      \\ \midrule[0.8pt]
ConStyle Restormer      & 34.33    & 31.71    & 28.51   & 34.37   & 31.74   & 28.52  & 34.89     & 32.66    & 29.64    & 35.01    & 32.74    & 29.71   \\
ConStyle v2 Restormer   & 34.33    & 31.71    & 28.51   & 34.37   & 31.74   & 28.52  & 34.89     & 32.67    & 29.66    & 35.01    & 32.77    & 29.71   \\
Differ.                 & {\color[HTML]{3166FF} 0}        & {\color[HTML]{3166FF} 0}        & {\color[HTML]{3166FF} 0}       & {\color[HTML]{3166FF} 0}       & {\color[HTML]{3166FF} +0}   & {\color[HTML]{3166FF} 0}      & {\color[HTML]{3166FF} 0}         & {\color[HTML]{3166FF} +0.01}    & {\color[HTML]{3166FF} +0.02}    & {\color[HTML]{3166FF} 0}        & {\color[HTML]{3166FF} +0.03}    & {\color[HTML]{3166FF} +0}   \\ \midrule[0.8pt]
ConStyle NAFNet         & 34.31    & 31.69    & 28.50   & 34.34   & 31.71   & 28.52  & 34.82     & 32.58    & 29.55    & 34.91    & 32.64    & 29.63   \\
ConStyle v2 NAFNet      & 34.33    & 31.71    & 28.52   & 34.36   & 31.73   & 28.54  & 34.88     & 32.66    & 29.65    & 34.96    & 32.72    & 29.68   \\
Differ.                 & {\color[HTML]{3166FF} +0.02}    & {\color[HTML]{3166FF} +0.02}    & {\color[HTML]{3166FF} +0.02}   & {\color[HTML]{3166FF} +0.02}   & {\color[HTML]{3166FF} +0.02}   & {\color[HTML]{3166FF} +0.02}  & {\color[HTML]{3166FF} +0.06}     & {\color[HTML]{3166FF} +0.08}    & {\color[HTML]{3166FF} +0.10}    & {\color[HTML]{3166FF} +0.05}    & {\color[HTML]{3166FF} +0.08}    & {\color[HTML]{3166FF} +0.05}   \\ \midrule[0.8pt]
ConStyle MAXIM-1S       & 34.25    & 31.63    & 28.43   & 34.28   & 31.65   & 28.44  & 34.56     & 32.26    & 29.10    & 34.64    & 32.32    & 29.13   \\
ConStyle v2 MAXIM-1S    & 34.19    & 31.55    & 28.34   & 34.22   & 31.58   & 28.36  & 34.60     & 32.32    & 29.19    & 34.70    & 32.40    & 29.26   \\
Differ.                 & {\color[HTML]{FE0000} -0.06}    & {\color[HTML]{FE0000} -0.08}    & {\color[HTML]{FE0000} -0.09}   & {\color[HTML]{FE0000} -0.06}   & {\color[HTML]{FE0000} -0.07}   & {\color[HTML]{FE0000} -0.08}  & {\color[HTML]{3166FF} +0.04}     & {\color[HTML]{3166FF} +0.06}    & {\color[HTML]{3166FF} +0.09}    & {\color[HTML]{3166FF} +0.06}    & {\color[HTML]{3166FF} +0.08}    & {\color[HTML]{3166FF} +0.13}   \\ \bottomrule[1.3pt]
\end{tabular}}
\label{denoising}
\end{table}

\subsection{Model Analyses}

\cref{comparaision} presents a comparison of parameters, computations, and speed between all models. All the results are obtained using input data of size (2,3,128,128), and the speed is the average of 10,000 inference. The reason why the parameters of the ConStyle v2/ConStyle models are fewer than the original models (except for Original Conv and ConStyle v2 Conv) is that, to demonstrate that the improvement of the ConStyle models is not brought by simply expanding the scale of the network, ConStyle models are downscaled by reducing the width and depth \cite{fan2024irconstyle}. Because of the introduction of the ConStyle part, the parameters of models will be increased by 1.19M.

\begin{table}[t]
\caption{Image motion deblurring and dehazing tested by PSNR/SSIM. Blue means better and red means worse.}
\scalebox{0.8}{
\begin{tabular}{cccccc}
\toprule[1.3pt]
Method                & GoPro \cite{gopro}         & RealBlur-R \cite{realblur}    & RealBlur-J \cite{realblur}    & HIDE \cite{hide}          & SOTS outdoor \cite{ots_sots}  \\ \midrule[0.8pt]
ConStyle Restormer    & 31.45/0.9208  & 33.94/0.9454  & 26.63/0.8288  & 30.20/0.9098  & 30.85/0.9760  \\
ConStyle v2 Restormer & 31.36/0.9200  & 33.95/0.9447  & 26.63/0.8298  & 30.10/0.9080  & 31.32/0.9789  \\
Differ.               & {\color[HTML]{FE0000} -0.09}/{\color[HTML]{FE0000} -0.0008} & {\color[HTML]{3166FF} +0.01}/{\color[HTML]{FE0000} -0.0007} & {\color[HTML]{3166FF} 0}/{\color[HTML]{FE0000} -0.0010}     & {\color[HTML]{FE0000} -0.10}/{\color[HTML]{FE0000} -0.0018} & {\color[HTML]{3166FF} +0.47}/{\color[HTML]{3166FF} +0.0029} \\ \midrule[0.8pt]
ConStyle NAFNet       & 31.56/0.9230  & 33.89/0.9441  & 26.61/0.8308  & 30.24/0.9106  & 30.73/0.9743  \\
ConStyle v2 NAFNet    & 31.68/0.9254  & 33.90/0.9444  & 26.63/0.8311  & 30.32/0.9115  & 32.34/0.9812  \\
Differ.               & {\color[HTML]{3166FF} +0.12}/{\color[HTML]{3166FF} +0.0024} & {\color[HTML]{3166FF} +0.01}/{\color[HTML]{3166FF} +0.0003} & {\color[HTML]{3166FF} +0.02}/{\color[HTML]{3166FF} +0.0003} & {\color[HTML]{3166FF} +0.08}/{\color[HTML]{3166FF} +0.0009} & {\color[HTML]{3166FF} +1.61}/{\color[HTML]{3166FF} +0.0069} \\ \midrule[0.8pt]
ConStyle MAXIM-1S     & 30.77/0.9093  & 33.93/0.9435  & 26.54/0.8261  & 29.26/0.8951  & 30.59/0.9713  \\
ConStyle v2 MAXIM-1S  & 31.01/0.9159  & 33.79/0.9407  & 26.47/0.8248  & 29.36/0.8952  & 30.68/0.9751  \\
Differ.               & {\color[HTML]{3166FF} +0.24}/{\color[HTML]{3166FF} +0.0066} & {\color[HTML]{FE0000} -0.14}/{\color[HTML]{FE0000} -0.0028} & {\color[HTML]{FE0000} -0.07}/{\color[HTML]{FE0000} -0.0013} & {\color[HTML]{3166FF} +0.10}/{\color[HTML]{3166FF} +0.0001} & {\color[HTML]{3166FF} +0.09}/{\color[HTML]{3166FF} +0.0038} \\ \bottomrule[1.3pt]
\end{tabular}}
\label{deblurring}
\end{table}

\subsection{All-in-One Image Restoration Result}
To verify the overall performance of our methods, we calculate the average PSNR/SSIM on 27 benchmarks. As shown in \cref{averge_performance}, except that the performance of Pre-train Conv is lower than that of Original Conv and ConStyle Conv on PSNR, the performance of other Pre-train, Class, and ConStyle v2 models are significantly higher than that of Original and ConStyle models. This highlights the effectiveness of three proposed methods: unsupervised pre-training, using pretext task, and using knowledge distillation. Due to space constraints, the detail results of Restormer, NAFNet, MAXIM-1S, and Conv models can be found in the \textbf{supplementary material}. While the ConStyle v2 models may not outperform ConStyle, Pre-train, and Class models in certain benchmarks, they still show significant improvement over the Original models. It is worth noting that scaling up the ConStyle v2 models to the size of the Original models could potentially yield even better results.

\subsection{Single Image Restoration Result}
Considering the significant improvement demonstrated by ConStyle v2 in handling multiple degradations, it is worth investigating whether this method can also enhance general restoration models to address specific degradations. We simply utilize the same training settings as IRConStyle \cite{fan2024irconstyle} for specific degradation scenarios. A comparison between ConStyle v2 models and ConStyle models is conducted across motion deblurring, denoising, and dehazing tasks. The denoising results are presented in \cref{denoising}, while the results for motion deblurring and dehazing are shown \cref{deblurring}.

\textbf{For denoising}, except for the performance of ConStyle v2 MAXIM-1S slightly worse than ConStyle MAXIM-1S on CBSD68, the overall performance of ConStyle v2 models is better than ConStyle models. \textbf{For dehazing}, ConStyle v2 models significantly outperform ConStyle models, even by 1.61 dB on NAFNet models. \textbf{For motion deblurring}, ConStyle v2 NAFNet is superior to ConStyle NAFNet but is indistinguishable from ConStyle on Restormer and MAXIM-1S. \textbf{In general}, 
for specific degradation, ConStyle v2 models do not require an accurate number of iterations to freeze the weight, while ConStyle models need to do so to avoid the problem of model collapse. Therefore, ConStyle v2 makes the entire IRConStyle framework more efficient.

\subsection{All-in-One Image Restoration Visual Result}
\begin{figure}[tb]
\centering
\includegraphics[width=\linewidth]{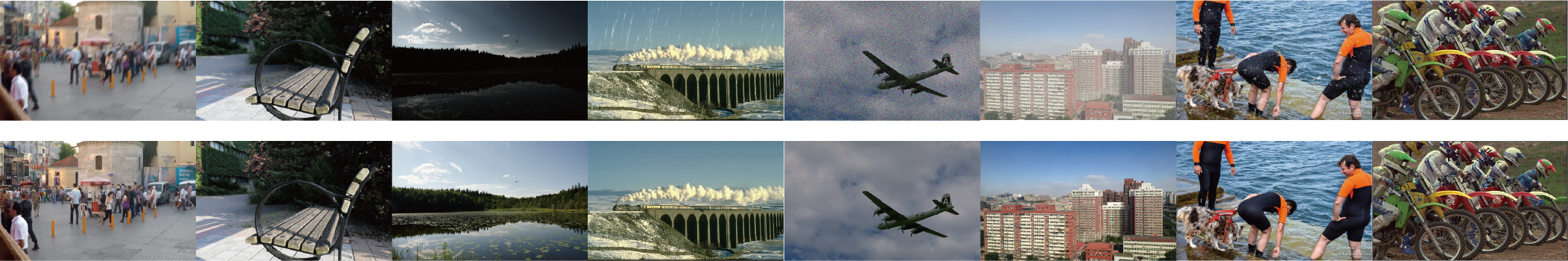}
\caption{The origin degraded images (top) and target images (bottom).}     
\label{target}
\end{figure}

\begin{figure}[tb]
\centering
\includegraphics[width=\linewidth]{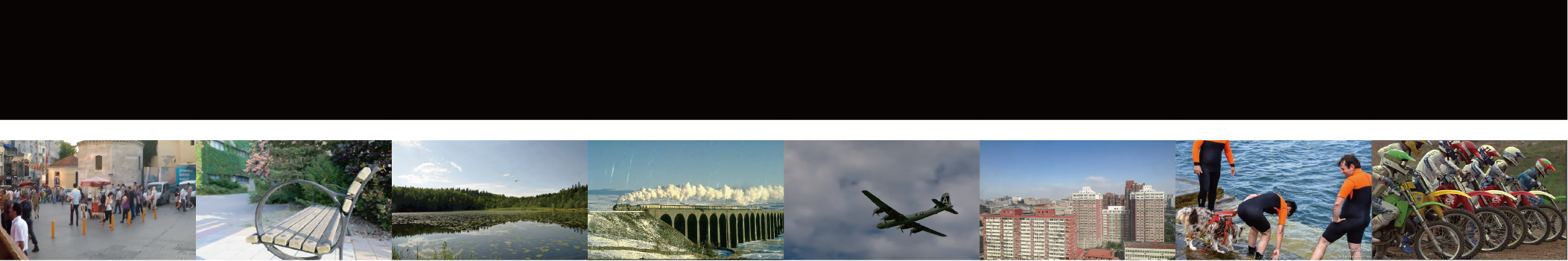}
\caption{The visual results of the Restormer (top) and ConStyle v2 Restormer (bottom).}  
\label{restormer}
\end{figure}

\begin{figure}[tb]
\centering
\includegraphics[width=\linewidth]{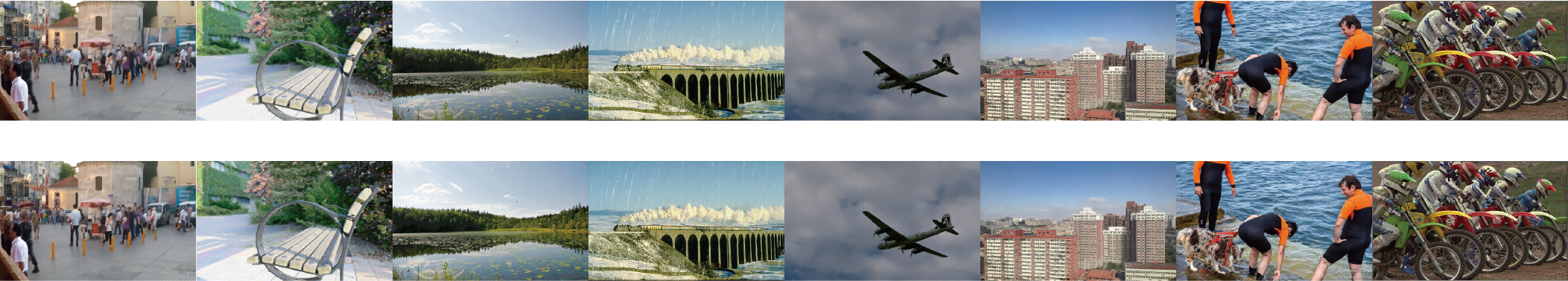}
\caption{The visual results of the NAFNet (top) and ConStyle v2 NAFNet (bottom).}  
\label{nafnet}
\end{figure}

\begin{figure}[tb]
\centering
\includegraphics[width=\linewidth]{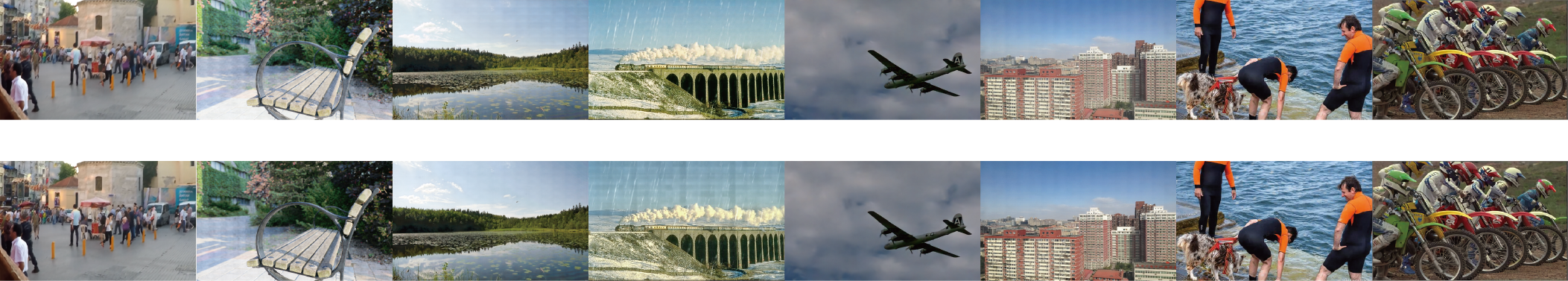}
\caption{The visual results of the MAXIM-1S (top) and ConStyle v2 MAXIM-1S (bottom).}   
\label{maxim}
\end{figure}

\begin{figure}[tb]
\centering
\includegraphics[width=\linewidth]{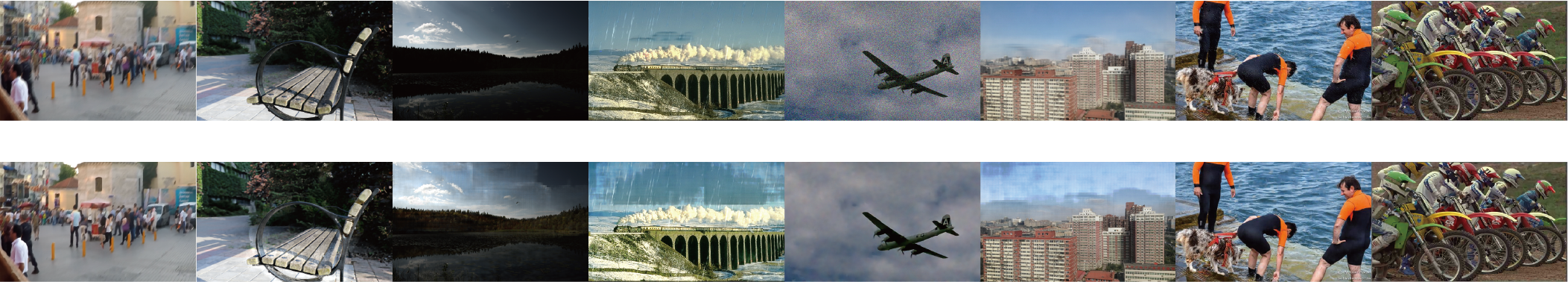}
\caption{The visual results of the Original Conv (top) and ConStyle v2 Conv (bottom).}  
\label{conv}
\end{figure}

We present the visual results of the Original models and ConStyle v2 models on the GoPro, DPDD, LoL v2, CBSD68, Rain100L, SOTS outdoors, snow100K-M, and LIVE1. Due to the space constraint and fair comparasions, for all tasks, we only select one identical image for each model. The origin degrdaded and target images are shown in \cref{target}. The visual results of the Restormer and ConStyle v2 Restormer are shown in \cref{restormer}, the visual results of the NAFNet and ConStyle v2 NAFNet are shown in \cref{nafnet}, the visual results of the MAXIM-1S and ConStyle v2 MAXIM-1S are shown in \cref{maxim}, and the visual results of the Original Conv and ConStyle v2 Conv are shown in \cref{conv}.

\subsection{Ablation Studies}
In the process of improving ConStyle to ConStyle v2, the results of optimization are obtained step by step (\cref{sec:constylev2}): unsupervised pre-training, adding a pretext task, and adopting knowledge distillation. Therefore, our whole improvement process is also the process of ablation studies. In addition, for every step of improvement, we conduct all models on the Mix Degradations datasets for fair comparisons (see supplemental materials for details).

\section{Conclusions and Limitations}
\label{sec:conclusion}

\subsection{Conclusions}
This paper leverages the unsupervised pre-training, pretext task, and knowledge distillation to improve ConStyle into a strong prompter for all-in-one image restoration. ConStyle v2 not only significantly improves the performance of the Original models under multiple degradations settings, but also solves the issue of model collapse (observed in Restormer) and unstable training caused by model limitations (observed in Original Conv). Moreover, the redundant operations that manually select specific iterations to freeze weights across different models and tasks in ConStyle are avoided, and the performance of ConStyle v2 models under certain specific degradations is also improved. Finally, due to the lack of training datasets for multiple degradations, the Mix Degradations datasets is collected and introduced.

\subsection{Limitations}
\label{limitation}
Despite many advantages as described in the conclusion and experiments, there are inevitably two limitations. Firstly, ConStyle v2 exhibits limited improvements in low-light, deraining, and defocus deblurring tasks compared to other tasks. This is evident in the results of ConStyle v2 Conv on LoL v1 and ConStyle v2 MAXIM-1S on DPDD. The reason is that during the pre-training stage, the generation methods of rain, low light, and defocus blur are not included in the degradation generation (Real-ESRGAN and ImageNet-C) since the effective method of the synthetic defocus blur, rain, and low light is still a challenge in the IR community.  Secondly, while ConStyle v2 models have shown promising results in quantifiable indicators such as PSNR/SSIM, the visual improvements in ConStyle v2 Conv seem to be less pronounced. This may be attributed to the inherent limitations of the Original Conv model, which is not specifically tailored for image restoration. However, ConStyle v2 demonstrates significant visual enhancements in most tasks on MAXIM-1S and NAFNet.

\bibliographystyle{splncs04}
\bibliography{egbib}

\appendix
\title{Supplementary Material: A Strong Prompter for All-in-One Image Restoration}

\titlerunning{Abbreviated paper title}

\author{Dongqi Fan \and
Junhao Zhang \and Liang Chang}

\authorrunning{Fan et al.}

\institute{University of Electronic Science and Technology of China, Chengdu, CN
\\
\email{\{dongqifan, junhaozhang\}@std.uestc.edu.cn} \\ \email{liangchang@uestc.edu.cn} 
}

\maketitle

\section{All-in-One Image Restoration Result}

\begin{table}[htbp]
\centering
\caption{Image motion deblurring, deraining, low-light enhancement, defocus deblurring, and dehazing of \textbf{Restormer} models tested by PSNR/SSIM. Red means the best and blue means the second best.}
\scalebox{0.95}{
\begin{tabular}{cccccc}
\toprule[1.3pt]
Datasets     & \makebox[0.17\textwidth][c]{Original}     & \makebox[0.17\textwidth][c]{ConStyle}     & \makebox[0.17\textwidth][c]{Pre-train}    & \makebox[0.17\textwidth][c]{Class}        & \makebox[0.17\textwidth][c]{ConStyle v2}    \\ \midrule[0.8pt]
GoPro \cite{gopro}      & 5.91/0.0062  & 25.64/0.7912 & 27.94/0.8558 & 27.93/0.8554 & {\color[HTML]{FE0000} 28.03}/{\color[HTML]{FE0000} 0.8585} \\
HIDE \cite{hide}       & 5.99/0.0157  & 23.87/0.7560 & 26.08/0.8338 & 26.51/0.8372 & {\color[HTML]{FE0000} 26.53}/{\color[HTML]{FE0000} 0.8393} \\
RealBlur-J \cite{realblur} & 10.97/0.0591 & 18.49/0.5820 & 18.50/0.4209 & 19.79/0.6579 & {\color[HTML]{3166FF} 19.39}/0.4664 \\
RealBlur-R \cite{realblur} & 8.56/0.3099  & 17.23/0.6388 & 19.88/0.4953 & 17.69/0.4213 & {\color[HTML]{FE0000} 23.23}/{\color[HTML]{FE0000} 0.7408} \\ \midrule[0.8pt]
Rain100H \cite{rain100l}   & 6.25/0.0032  & 18.41/0.5615 & 24.95/0.7818 & 25.23/0.7852 & {\color[HTML]{FE0000} 25.54}/{\color[HTML]{FE0000} 0.7949} \\
Rain100L \cite{rain100l}   & 6.25/0.0032  & 24.33/0.8136 & 22.04/0.8340 & 23.73/0.8503 & {\color[HTML]{FE0000} 24.49}/{\color[HTML]{3166FF} 0.8358} \\
Test1200 \cite{test1200zhang2018density}   & 6.33/0.0169  & 27.72/0.8151 & 31.00/0.8847 & 31.30/0.8888 & {\color[HTML]{FE0000} 31.41}/{\color[HTML]{FE0000} 0.8926} \\
Test2800 \cite{test2800fu2017removing}   & 6.29/0.0058  & 27.40/0.8353 & 30.93/0.9111 & 30.99/0.9119 & {\color[HTML]{FE0000} 31.06}/{\color[HTML]{FE0000} 0.9130} \\ \midrule[0.8pt]
FiveK \cite{fivekbychkovsky2011learning}      & 6.64/0.0307  & 18.39/0.8026 & 24.50/0.9062 & 24.59/0.9058 & {\color[HTML]{3166FF} 24.46}/{\color[HTML]{3166FF} 0.9058} \\
LoL v1 \cite{lolv1wei1808deep}     & 6.38/0.0177  & 18.54/0.7231 & 21.35/0.8003 & 21.87/0.8015 & {\color[HTML]{3166FF} 21.75}/{\color[HTML]{FE0000} 0.8189} \\
LoL v2 \cite{lolv2yang2021sparse}     & 5.57/0.0145  & 15.56/0.7042 & 24.06/0.9145 & 24.46/0.9165 & {\color[HTML]{3166FF} 24.10}/0.9136 \\ \midrule[0.8pt]
DPDD \cite{dpddabuolaim2020defocus}       & 7.73/0.0143  & 20.22/0.7037 & 15.00/0.6134 & 15.71/0.6104 & 14.38/0.5988 \\ \midrule[0.8pt]
SOTS \cite{ots_sots}       & 5.23/0.0065  & 24.80/0.9326 & 28.64/0.9694 & 28.86/0.9700 & {\color[HTML]{FE0000} 29.86}/{\color[HTML]{FE0000} 0.9704} \\ \bottomrule[1.3pt]
\end{tabular}}
\label{restormer_results}
\end{table}

\begin{table}[htbp]
\centering
\caption{Image denoising, JPEG artifact removal, and desnowing of \textbf{Restormer} models tested by PSNR/SSIM. Red means the best and blue means the second best.}
\scalebox{0.95}{
\begin{tabular}{ccccccc}
\toprule[1.3pt]
\multicolumn{2}{c}{Datasets}   & \makebox[0.17\textwidth][c]{Original}  & \makebox[0.17\textwidth][c]{ConStyle}  & \makebox[0.17\textwidth][c]{Pre-train}  & \makebox[0.17\textwidth][c]{Class}  & \makebox[0.17\textwidth][c]{ConStyle v2}   \\ \midrule[0.8pt]
                          & 15 & 6.62/0.0044        & 31.99/0.8844       & 33.33/0.9100        & 33.35/0.9117    & {\color[HTML]{FE0000} 33.40}/{\color[HTML]{FE0000} 0.9130}          \\
CBSD68 \cite{bsd68}                    & 25 & 6.62/0.0044        & 29.33/0.8146       & 30.40/0.8517        & 30.39/0.8515    & {\color[HTML]{FE0000} 30.45}/{\color[HTML]{FE0000} 0.8519}          \\
                          & 50 & 6.62/0.0044        & 25.31/0.6445       & 25.63/0.7035        & 25.77/0.7081    & {\color[HTML]{FE0000} 25.79}/{\color[HTML]{FE0000} 0.7098}          \\ \midrule[0.8pt]
\multirow{3}{*}{Urban100 \cite{urban100}} & 15 & 5.72/0.218         & 31.15/0.8923       & 32.43/0.9107        & 32.77/0.9100    & {\color[HTML]{3166ff} 32.45}/{\color[HTML]{3166ff} 0.9106}          \\
                          & 25 & 5.72/0.218         & 28.49/0.8351       & 29.60/0.8602        & 29.86/0.8661    & {\color[HTML]{3166ff} 29.65}/{\color[HTML]{3166ff} 0.8619}          \\
                          & 50 & 5.72/0.218         & 24.44/0.6716       & 24.69/0.7206        & 24.97/0.7366    & {\color[HTML]{FE0000} 25.07}/{\color[HTML]{FE0000} 0.7416}          \\ \midrule[0.8pt]
\multirow{4}{*}{LIVE1 \cite{livesheikh2005live}}    & 10 & 6.20/0.0020        & 25.72/0.7512       & 27.00/0.7839        & 27.05/0.7852    & {\color[HTML]{FE0000} 27.12}/{\color[HTML]{FE0000} 0.7887}            \\
                          & 20 & 6.20/0.0020        & 28.09/0.8329       & 29.37/0.8558        & 29.42/0.8570    & {\color[HTML]{FE0000} 29.47}/{\color[HTML]{FE0000} 0.8582}          \\
                          & 30 & 6.20/0.0020        & 29.29/0.8674       & 30.67/0.8863        & 30.73/0.8874    & {\color[HTML]{FE0000} 30.77}/{\color[HTML]{FE0000} 0.8882}         \\
                          & 40 & 6.20/0.0020        & 30.04/0.8866       & 31.58/0.9034        & 31.63/0.9044    & {\color[HTML]{FE0000} 31.67}/{\color[HTML]{FE0000} 0.9050}          \\ \midrule[0.8pt]
\multirow{3}{*}{Snow100K \cite{12liu2018desnownet}} & S  & 6.43/0.0203        & 29.12/0.8778       & 33.90/0.9418        & 34.04/0.9430    & {\color[HTML]{FE0000} 34.23}/{\color[HTML]{FE0000} 0.9443}        \\
                          & M  & 6.44/0.0203        & 26.17/0.8549       & 32.17/0.9300        & 32.36/0.9316    & {\color[HTML]{FE0000} 32.49}/{\color[HTML]{FE0000} 0.9330 }         \\
                          & L  & 6.45/0.0214        & 23.57/0.7808       & 28.28/0.8761        & 28.46/0.8793    & {\color[HTML]{FE0000} 28.50}/{\color[HTML]{FE0000} 0.8809 }        \\ \midrule[0.8pt]
\multicolumn{2}{c}{CSD \cite{12liu2018desnownet}}        & 5.43/0.0078        & 16.03/0.7596       & 16.00/0.7404        & 15.35/0.7595    & {\color[HTML]{FE0000} 16.13}/{\color[HTML]{FE0000} 0.7625 }          \\ \bottomrule[1.3pt]
\end{tabular}}
\label{restormer_results2}
\end{table}

\begin{table}[htbp]
\centering
\caption{Image motion deblurring, deraining, low-light enhancement, defocus deblurring, and dehazing of \textbf{NAFNet} models tested by PSNR/SSIM. Red means the best and blue means the second best.}
\scalebox{0.95}{
\begin{tabular}{cccccc}
\toprule[1.3pt]
Datasets     & \makebox[0.17\textwidth][c]{Original}     & \makebox[0.17\textwidth][c]{ConStyle}     & \makebox[0.17\textwidth][c]{Pre-train}    & \makebox[0.17\textwidth][c]{Class}        & \makebox[0.17\textwidth][c]{ConStyle v2}    \\ \midrule[0.8pt]
GoPro \cite{gopro}     & 27.52/0.8461    & 27.81/0.5839    & 27.81/0.8536     & 27.80/0.8530 & {\color[HTML]{FE0000} 27.82}/{\color[HTML]{FE0000} 0.8540}       \\
HIDE \cite{hide}       & 25.01/0.8173    & 25.55/0.8293    & 25.27/0.8228     & 25.30/0.8125 & {\color[HTML]{3166ff} 25.38}/{\color[HTML]{3166ff} 0.8243}       \\
RealBlur-J \cite{realblur} & 23.11/0.7418    & 24.32/0.7688    & 23.67/0.7435     & 25.17/0.7918 & 23.64/0.7419       \\
RealBlur-R \cite{realblur} & 29.18/0.7915    & 27.72/0.7397    & 28.63/0.7596     & 27.49/0.7281 & 27.34/0.7272       \\ \midrule[0.8pt]
Rain100H \cite{rain100l}   & 24.47/0.7666    & 21.93/0.7943    & 24.53/0.7437     & 24.72/0.7748 & {\color[HTML]{3166ff} 25.00}/{\color[HTML]{3166ff} 0.7777}       \\
Rain100L \cite{rain100l}   & 21.93/0.7943    & 22.03/0.8003    & 21.85/0.8110     & 21.09/0.7958 & {\color[HTML]{FE0000} 22.12}/{\color[HTML]{FE0000} 0.8111}       \\
Test1200 \cite{test1200zhang2018density}   & 30.68/0.8820    & 29.38/0.8631    & 29.40/0.8793     & 29.41/0.8608 & {\color[HTML]{FE0000} 29.62}/{\color[HTML]{3166ff} 0.8669}       \\
Test2800 \cite{test2800fu2017removing}  & 30.70/0.9020    & 30.74/0.9087    & 30.97/0.9017     & 30.73/0.9087 & {\color[HTML]{3166ff} 30.75}/{\color[HTML]{FE0000} 0.9089}       \\ \midrule[0.8pt]
FiveK \cite{fivekbychkovsky2011learning}      & 24.43/0.9022    & 24.41/0.9028    & 24.42/0.9030     & 24.45/0.9018 & {\color[HTML]{FE0000} 24.47}/{\color[HTML]{FE0000} 0.9038}       \\
LoL v1 \cite{lolv1wei1808deep}     & 21.73/0.7878    & 21.89/0.7804    & 21.74/0.8022     & 21.96/0.7993 & {\color[HTML]{FE0000} 22.05}/{\color[HTML]{FE0000} 0.7995}       \\
LoL v2 \cite{lolv2yang2021sparse}     & 24.08/0.9087    & 23.77/0.9085    & 24.08/0.9094     & 22.05/0.7995 & {\color[HTML]{FE0000} 24.47}/{\color[HTML]{FE0000} 0.9128}       \\ \midrule[0.8pt]
DPDD \cite{dpddabuolaim2020defocus}       & 13.34/0.5684    & 14.09/0.5860    & 14.11/0.5689     & 14.22/0.5817 & {\color[HTML]{FE0000} 14.25}/{\color[HTML]{FE0000} 0.5961}       \\ \midrule[0.8pt]
SOTS \cite{ots_sots}       & 29.01/0.9452    & 29.18/0.9650    & 28.51/0.9213     & 29.13/0.9646 & {\color[HTML]{FE0000} 29.19}/{\color[HTML]{FE0000} 0.9651}       \\ \bottomrule[1.3pt]
\end{tabular}}
\end{table}

\begin{table}[htbp]
\centering
\caption{Image denoising, JPEG artifact removal, and desnowing of \textbf{NAFNet} models tested by PSNR/SSIM. Red means the best and blue means the second best.}
\scalebox{0.95}{
\begin{tabular}{ccccccc}
\toprule[1.3pt]
\multicolumn{2}{c}{Datasets}   & \makebox[0.17\textwidth][c]{Original}  & \makebox[0.17\textwidth][c]{ConStyle}  & \makebox[0.17\textwidth][c]{Pre-train}  & \makebox[0.17\textwidth][c]{Class}  & \makebox[0.17\textwidth][c]{ConStyle v2}   \\ \midrule[0.8pt]
                          & 15 & 33.00/0.9004    & 33.20/0.9098    & 33.19/0.9005     & 33.20/0.9011 & {\color[HTML]{FE0000} 33.22}/{\color[HTML]{FE0000} 0.9099}       \\
CBSD68 \cite{bsd68}                    & 25 & 30.19/0.8417    & 29.94/0.8416    & 30.00/0.8357     & 30.22/0.8427 & {\color[HTML]{FE0000} 30.27}/{\color[HTML]{FE0000} 0.8440}       \\
                          & 50 & 25.13/0.6917    & 22.64/0.6433    & 25.34/0.6963     & 25.24/0.6994 & {\color[HTML]{FE0000} 25.41}/{\color[HTML]{FE0000} 0.7058}       \\ \midrule[0.8pt]
\multirow{3}{*}{Urban100 \cite{urban100}} & 15 & 32.81/0.9226    & 29.77/0.8670    & 32.90/0.9242     & 32.83/0.9241 & 32.82/0.9237       \\
                          & 25 & 29.83/0.8718    & 25.20/0.7623    & 28.90/0.8655     & 29.91/0.8762 & {\color[HTML]{3166ff} 29.90}/{\color[HTML]{3166ff} 0.8761}       \\
                          & 50 & 24.28/0.7247    & 16.17/0.4767    & 24.38/0.7425     & 24.36/0.7459 & {\color[HTML]{FE0000} 24.49}/{\color[HTML]{FE0000} 0.7496}       \\ \midrule[0.8pt]
\multirow{4}{*}{LIVE1 \cite{livesheikh2005live}}    & 10 & 26.94/0.7828    & 26.96/0.7824    & 26.85/0.7802     & 26.96/0.7829 & {\color[HTML]{FE0000} 26.99}/{\color[HTML]{FE0000} 0.7842}       \\
                          & 20 & 29.28/0.8544    & 29.29/0.8539    & 29.28/0.8526     & 29.30/0.8545 & {\color[HTML]{FE0000} 29.31}/{\color[HTML]{FE0000} 0.8547}       \\
                          & 30 & 30.58/0.8840    & 30.59/0.8850    & 30.49/0.8833     & 30.60/0.8851 & {\color[HTML]{FE0000} 30.60}/{\color[HTML]{3166ff} 0.8850}       \\
                          & 40 & 31.48/0.9023    & 31.50/0.9027    & 31.54/0.9028     & 31.49/0.9025 & {\color[HTML]{3166ff} 31.50}/0.9026       \\ \midrule[0.8pt]
\multirow{3}{*}{Snow100K \cite{12liu2018desnownet}} & S  & 33.41/0.9399    & 33.45/0.9399    & 33.40/0.9323     & 33.38/0.9397 & {\color[HTML]{FE0000} 33.51}/{\color[HTML]{FE0000} 0.9404}       \\
                          & M  & 31.77/0.9276    & 31.77/0.9275    & 31.65/0.9206     & 31.72/0.9272 & {\color[HTML]{FE0000} 31.79}/{\color[HTML]{FE0000}0.9280}       \\
                          & L  & 27.60/0.8710    & 27.64/0.8703    & 28.30/0.8771     & 27.59/0.8697 & {\color[HTML]{3166ff} 27.68}/{\color[HTML]{3166ff}0.8711}       \\ \midrule[0.8pt]
\multicolumn{2}{c}{CSD \cite{10chen2021all}}        & 16.49/0.7493    & 16.83/0.7545    & 16.69/0.7606     & 17.69/0.7677 &  {\color[HTML]{3166ff} 17.29}/{\color[HTML]{3166ff}0.7621}                  \\ \bottomrule[1.3pt]
\end{tabular}}
\end{table}

\begin{table}[htbp]
\centering
\caption{Image motion deblurring, deraining, low-light enhancement, defocus deblurring, and dehazing of \textbf{MAXIM-1S} models tested by PSNR/SSIM. Red means the best and blue means the second best.}
\scalebox{0.95}{
\begin{tabular}{cccccc}
\toprule[1.3pt]
Datasets     & \makebox[0.17\textwidth][c]{Original}     & \makebox[0.17\textwidth][c]{ConStyle}     & \makebox[0.17\textwidth][c]{Pre-train}    & \makebox[0.17\textwidth][c]{Class}        & \makebox[0.17\textwidth][c]{ConStyle v2}    \\ \midrule[0.8pt]
GoPro \cite{gopro}      & 26.99/0.8294      & 27.35/0.8319      & 27.28/0.8379       & 27.35/0.8391   & {\color[HTML]{FE0000} 27.37}/{\color[HTML]{FE0000}0.8401}         \\
HIDE \cite{hide}       & 23.92/0.7764      & 25.13/0.8079      & 24.28/0.7924       & 25.54/0.8107   & {\color[HTML]{FE0000} 25.35}/{\color[HTML]{FE0000}0.8096}         \\
RealBlur-J \cite{realblur} & 24.05/0.7673      & 24.37/0.7574      & 23.33/0.7278       & 24.74/0.7669   & {\color[HTML]{FE0000} 25.05}/{\color[HTML]{FE0000}0.7751}         \\
RealBlur-R \cite{realblur} & 23.11/0.7305      & 24.94/0.6566      & 22.22/0.5779       & 25.83/0.6676   & {\color[HTML]{FE0000} 27.65}/{\color[HTML]{FE0000}0.7341}        \\ \midrule[0.8pt]
Rain100H \cite{rain100l}    & 21.58/0.7020      & 22.16/0.7777      & 22.84/0.7544       & 22.82/0.7503   & {\color[HTML]{FE0000} 23.75}/{\color[HTML]{3166ff}0.7603}         \\
Rain100L \cite{rain100l}    & 21.58/0.8020      & 21.81/0.7961      & 20.74/0.7860       & 19.90/0.7693   & {\color[HTML]{FE0000} 22.64}/{\color[HTML]{FE0000}0.8076}        \\
Test1200 \cite{test1200zhang2018density}   & 30.52/0.8826      & 29.04/0.8783      & 30.81/0.8839       & 30.48/0.8815   & {\color[HTML]{3166ff} 30.76}/{\color[HTML]{3166ff}0.8829}        \\
Test2800 \cite{test2800fu2017removing}  & 30.50/0.9060      & 28.34/0.8675      & 30.70/0.8999       & 30.83/0.9109   & 30.55/{\color[HTML]{3166ff}0.9075}         \\ \midrule[0.8pt]
FiveK \cite{fivekbychkovsky2011learning}       & 23.94/0.8953      & 24.10/0.8907      & 24.16/0.8990       & 24.15/0.9008   & {\color[HTML]{FE0000} 24.17}/{\color[HTML]{FE0000}0.8993}         \\
LoL v1 \cite{lolv1wei1808deep}     & 20.99/0.8000      & 20.97/0.8119      & 20.88/0.8027       & 21.03/0.8078   & {\color[HTML]{3166ff} 21.59}/{\color[HTML]{3166ff}0.8089}         \\
LoL v2 \cite{lolv2yang2021sparse}     & 23.26/0.9036      & 24.09/0.9111      & 24.13/0.9055       & 24.44/0.9122   & {\color[HTML]{3166ff} 24.40}/{\color[HTML]{3166ff}0.9112}             \\ \midrule[0.8pt]
DPDD \cite{dpddabuolaim2020defocus}       & 14.71/0.5969      & 15.09/0.6082      & 14.84/0.5982       & 15.47/0.6079   & 14.81/0.5989         \\ \midrule[0.8pt]
SOTS \cite{ots_sots}       & 28.55/0.9586      & 29.36/0.9651      & 29.34/0.9660       & 29.24/0.9600   & {\color[HTML]{FE0000} 29.47}/{\color[HTML]{FE0000}0.9679}         \\ \bottomrule[1.3pt]
\end{tabular}}
\end{table}

\begin{table}[htbp]
\centering
\caption{Image denoising, JPEG artifact removal, and desnowing of \textbf{MAXIM-1S} models tested by PSNR/SSIM. Red means the best and blue means the second best.}
\scalebox{0.95}{
\begin{tabular}{ccccccc}
\toprule[1.3pt]
\multicolumn{2}{c}{Datasets}   & \makebox[0.17\textwidth][c]{Original}  & \makebox[0.17\textwidth][c]{ConStyle}  & \makebox[0.17\textwidth][c]{Pre-train}  & \makebox[0.17\textwidth][c]{Class}  & \makebox[0.17\textwidth][c]{ConStyle v2}   \\ \midrule[0.8pt]
                          & 15 & 33.11/0.9094   & 33.20/0.9111   & 33.24/0.9113    & 33.29/0.9128 & {\color[HTML]{3166ff} 33.22}/{\color[HTML]{3166ff}0.9113}      \\
CBSD68 \cite{bsd68}                    & 25 & 30.13/0.8462   & 30.27/0.8469   & 30.25/0.8470    & 30.33/0.8486 & {\color[HTML]{3166ff} 30.27}/{\color[HTML]{3166ff}0.8477}      \\
                          & 50 & 25.43/0.7102   & 25.50/0.7172   & 25.53/0.7195    & 25.67/0.7228 & {\color[HTML]{3166ff} 25.63}/0.7132      \\ \midrule[0.8pt]
\multirow{3}{*}{Urban100 \cite{urban100}} & 15 & 32.47/0.9190   & 32.11/0.9186   & 32.66/0.9202    & 32.71/0.9187 & {\color[HTML]{FE0000} 32.77}/{\color[HTML]{FE0000}0.9238}      \\
                          & 25 & 29.50/0.8693   & 29.04/0.8672   & 29.83/0.8689    & 29.86/0.8770 & {\color[HTML]{FE0000} 29.86}/{\color[HTML]{FE0000}0.8777}      \\
                          & 50 & 24.54/0.7499   & 24.07/0.7456   & 25.01/0.7689    & 25.07/0.7718 & 24.94/0.7629      \\ \midrule[0.8pt]
\multirow{4}{*}{LIVE1 \cite{livesheikh2005live}}    & 10 & 26.82/0.7824   & 24.93/0.7519   & 26.82/0.7841    & 26.89/0.7830 & {\color[HTML]{FE0000} 26.90}/{\color[HTML]{FE0000}0.7854}      \\
                          & 20 & 29.06/0.8504   & 27.09/0.8345   & 29.24/0.8462    & 29.18/0.8506 & {\color[HTML]{FE0000} 29.27}/{\color[HTML]{FE0000}0.8564}      \\
                          & 30 & 30.19/0.8778   & 28.40/0.8683   & 30.45/0.8866    & 30.68/0.8876 & {\color[HTML]{3166ff} 30.59}/{\color[HTML]{3166ff}0.8868}      \\
                          & 40 & 30.98/0.8941   & 28.73/0.8847   & 31.56/0.9038    & 31.58/0.9004 & 31.37/{\color[HTML]{3166ff}0.9012}      \\ \midrule[0.8pt]
\multirow{3}{*}{Snow100K \cite{12liu2018desnownet}} & S  & 33.05/0.9380   & 33.48/0.9324   & 33.38/0.9408    & 33.53/0.9406 & {\color[HTML]{FE0000} 33.54}/{\color[HTML]{3166ff}0.9420}      \\
                          & M  & 31.46/0.9261   & 31.86/0.9209   & 31.78/0.9292    & 31.81/0.9211 & {\color[HTML]{FE0000} 31.87}/{\color[HTML]{FE0000}0.9303}      \\
                          & L  & 27.78/0.8726   & 28.01/0.8760   & 27.95/0.8755    & 28.16/0.8791 & {\color[HTML]{3166ff} 28.03}/{\color[HTML]{3166ff}0.8773}     \\ \midrule[0.8pt]
\multicolumn{2}{c}{CSD \cite{10chen2021all}}        & 13.60/0.7048   & 13.93/0.7087   & 13.48/0.7335    & 14.70/0.7344 &         {\color[HTML]{3166ff} 14.27}/0.7147          \\ \bottomrule[1.3pt]
\end{tabular}}
\end{table}

\begin{table}[htbp]
\centering
\caption{Image motion deblurring, deraining, low-light enhancement, defocus deblurring, and dehazing of \textbf{Conv} models tested by PSNR/SSIM. Red means the best and blue means the second best.}
\scalebox{0.95}{
\begin{tabular}{clllll}
\toprule[1.3pt]
Datasets     & \makebox[0.17\textwidth][c]{Original}     & \makebox[0.17\textwidth][c]{ConStyle}     & \makebox[0.17\textwidth][c]{Pre-train}    & \makebox[0.17\textwidth][c]{Class}        & \makebox[0.17\textwidth][c]{ConStyle v2}    \\ \midrule[0.8pt]
GoPro \cite{gopro}      & 21.56/0.6900  & 24.61/0.7275  & 25.67/0.7922   & 26.35/0.8108 & {\color[HTML]{FE0000} 26.54}/{\color[HTML]{FE0000}0.8121}     \\
HIDE \cite{hide}       & 20.04/0.6523  & 23.18/0.7001  & 21.84/0.7303   & 23.17/0.7019 & {\color[HTML]{FE0000} 23.27}/{\color[HTML]{FE0000}0.7671}     \\
RealBlur-J \cite{realblur} & 23.55/0.6025  & 25.22/0.6936  & 25.30/0.7888   & 23.66/0.7621 & {\color[HTML]{FE0000} 26.10}/{\color[HTML]{FE0000}0.7979}     \\
RealBlur-R \cite{realblur} & 30.01/0.8375  & 32.02/0.8901  & 30.32/0.8874   & 30.80/0.8889 & {\color[HTML]{FE0000} 32.60}/{\color[HTML]{FE0000}0.9130}     \\ \midrule[0.8pt]
Rain100H \cite{rain100l}    & 10.11/0.3799  & 12.32/0.3374  & 15.30/0.5284   & 11.96/0.4334 & {\color[HTML]{3166ff} 12.40}/0.4236     \\
Rain100L \cite{rain100l}    & 17.64/0.6086  & 23.77/0.7262  & 18.89/0.7502   & 18.11/0.7265 & {\color[HTML]{3166ff} 18.95}/{\color[HTML]{3166ff}0.7312 }    \\
Test1200 \cite{test1200zhang2018density}   & 25.07/0.7105  & 21.27/0.6324  & 23.52/0.7902   & 28.41/0.8499 & {\color[HTML]{3166ff} 27.98}/{\color[HTML]{3166ff}0.8479}     \\
Test2800 \cite{test2800fu2017removing}  & 25.40/0.7675  & 21.88/0.9619  & 23.44/0.8105   & 29.65/0.8964 & {\color[HTML]{3166ff} 29.27}/{\color[HTML]{3166ff}0.8897}     \\ \midrule[0.8pt]
FiveK \cite{fivekbychkovsky2011learning}       & 17.19/0.7864  & 17.33/0.7005  & 17.72/0.7805   & 18.94/0.8032 & {\color[HTML]{FE0000} 19.50}/{\color[HTML]{FE0000}0.8337}    \\
LoL v1 \cite{lolv1wei1808deep}     & 7.77/0.1935   & 7.76/0.1887   & 8.32/0.2434    & 9.43/0.3634  & 7.96/0.2134      \\
LoL v2 \cite{lolv2yang2021sparse}     & 11.05/0.4467  & 10.93/0.4086  & 13.02/0.5814   & 13.03/0.6542 & {\color[HTML]{FE0000} 13.21}/{\color[HTML]{3166ff}0.6206}    \\ \midrule[0.8pt]
DPDD \cite{dpddabuolaim2020defocus}       & 20.32/0.5338  & 21.29/0.6129  & 19.88/0.6912   & 19.82/0.6825 & {\color[HTML]{FE0000} 21.92}/{\color[HTML]{FE0000}0.7143}     \\ \midrule[0.8pt]
SOTS \cite{ots_sots}       & 22.92/0.9063  & 16.25/0.7639  & 22.22/0.9065   & 21.49/0.9160 & {\color[HTML]{FE0000} 24.21}/{\color[HTML]{FE0000}0.9406}     \\ \bottomrule[1.3pt]
\end{tabular}}
\end{table}

\begin{table}[htbp]
\centering
\caption{Image denoising, JPEG artifact removal, and desnowing of \textbf{Conv} models tested by PSNR/SSIM. Red means the best and blue means the second best.}
\scalebox{0.95}{
\begin{tabular}{ccccccc}
\toprule[1.3pt]
\multicolumn{2}{c}{Datasets}   & \makebox[0.17\textwidth][c]{Original}  & \makebox[0.17\textwidth][c]{ConStyle}  & \makebox[0.17\textwidth][c]{Pre-train}  & \makebox[0.17\textwidth][c]{Class}  & \makebox[0.17\textwidth][c]{ConStyle v2}   \\ \midrule[0.8pt]
                          & 15 & 24.82/0.5730  & 24.56/0.5976  & 19.22/0.6107   & 32.92/0.9059 & {\color[HTML]{FE0000} 33.09}/{\color[HTML]{FE0000}0.9047}    \\
CBSD68 \cite{bsd68}                    & 25 & 20.54/0.3950  & 21.27/0.4416  & 17.22/0.4439   & 28.33/0.8345 & {\color[HTML]{FE0000} 30.12}/{\color[HTML]{FE0000}0.8354}    \\
                          & 50 & 15.01/0.1999  & 16.16/0.2436  & 15.77/0.2747   & 21.74/0.6674 & {\color[HTML]{FE0000} 25.16}/{\color[HTML]{FE0000}0.6830}    \\ \midrule[0.8pt]
\multirow{3}{*}{Urban100 \cite{urban100}} & 15 & 24.85/0.6123  & 23.25/0.6137  & 11.72/0.4582   & 29.34/0.8898 & {\color[HTML]{FE0000} 32.38}/{\color[HTML]{FE0000}0.9150}     \\
                          & 25 & 20.69/0.4569  & 20.48/0.4843  & 12.96/0.3982   & 25.31/0.8214 & {\color[HTML]{FE0000} 29.48}/{\color[HTML]{FE0000}0.8608}     \\
                          & 50 & 15.18/0.2676  & 15.84/0.3055  & 13.97/0.2979   & 18.43/0.6606 & {\color[HTML]{FE0000} 24.40}/{\color[HTML]{FE0000}0.7230}     \\ \midrule[0.8pt]
\multirow{4}{*}{LIVE1 \cite{livesheikh2005live}}    & 10 & 24.95/0.7396  & 23.59/0.6604  & 25.06/0.7449   & 26.93/0.7836 & {\color[HTML]{3166ff} 26.80}/{\color[HTML]{3166ff}0.7787}    \\
                          & 20 & 26.53/0.8191  & 25.03/0.7324  & 26.40/0.8170   & 29.30/0.8555 & {\color[HTML]{3166ff} 29.22}/{\color[HTML]{3166ff}0.8526}     \\
                          & 30 & 27.32/0.8542  & 25.69/0.7619  & 26.40/0.8408   & 30.61/0.8864 & {\color[HTML]{3166ff} 30.54}/{\color[HTML]{3166ff}0.8843 }     \\
                          & 40 & 27.83/0.8745  & 26.12/0.7793  & 26.40/0.8485   & 31.50/0.9035 & {\color[HTML]{3166ff} 31.40}/{\color[HTML]{3166ff}0.9017}    \\ \midrule[0.8pt]
\multirow{3}{*}{Snow100K \cite{12liu2018desnownet}} & S  & 26.70/0.8268  & 23.46/0.7341  & 24.72/0.8582   & 31.55/0.9129 & {\color[HTML]{FE0000} 31.85}/{\color[HTML]{FE0000}0.9243}     \\
                          & M  & 24.79/0.7978  & 21.97/0.7103  & 24.34/0.8427   & 31.14/0.9200 & {\color[HTML]{3166ff} 30.43}/{\color[HTML]{3166ff}0.9106}     \\
                          & L  & 22.22/0.6906  & 18.56/0.6144  & 23.09/0.7833   & 27.15/0.8589 & {\color[HTML]{3166ff} 26.12}/{\color[HTML]{3166ff}0.8456}     \\ \midrule[0.8pt]
\multicolumn{2}{c}{CSD \cite{10chen2021all}}        & 15.78/0.7259  & 14.56/0.6666  & 15.66/0.7367   & 13.66/0.6878 & {\color[HTML]{FE0000} 16.63}/{\color[HTML]{FE0000}0.7515}     \\ \bottomrule[1.3pt]
\end{tabular}}
\end{table}

\end{document}